# Predicting 3D shapes, masks, and properties of materials, liquids, and objects inside transparent containers, using the TransProteus CGI dataset

Sagi Eppel[*,1,2], Haoping Xu[2,3], Yi Ru Wang[5], Alan Aspuru-Guzik[*,1,2,3,4]

## Abstract

We present TransProteus, a dataset, and methods for predicting the 3D structure, masks, and properties of materials, liquids, and objects inside transparent vessels from a single image without prior knowledge of the image source and camera parameters. Manipulating materials in transparent containers is essential in many fields and depends heavily on vision. This work supplies a new procedurally generated dataset consisting of 50k images of liquids and solid objects inside transparent containers. The image annotations include 3D models, material properties (color/transparency/roughness...), and segmentation masks for the vessel and its content. The synthetic (CGI) part of the dataset was procedurally generated using 13k different objects, 500 different environments (HDRI), and 1450 material textures (PBR) combined with simulated liquids and procedurally generated vessels. In addition, we supply 104 real-world images of objects inside transparent vessels with depth maps of both the vessel and its content. We propose a camera agnostic method that predicts 3D models from an image as an XYZ map. This allows the trained net to predict the 3D model as a map with XYZ coordinates per pixel without prior knowledge of the image source. To calculate the training loss, we use the distance between pairs of points inside the 3D model instead of the absolute XYZ coordinates, which makes the loss function translation invariant. We use this to predict 3D models of vessels and their content from a single image. Finally, we demonstrate a network that uses a single image to predict the material properties of the vessel content and surface. Dataset is available at this URL. Code is available at this URL.

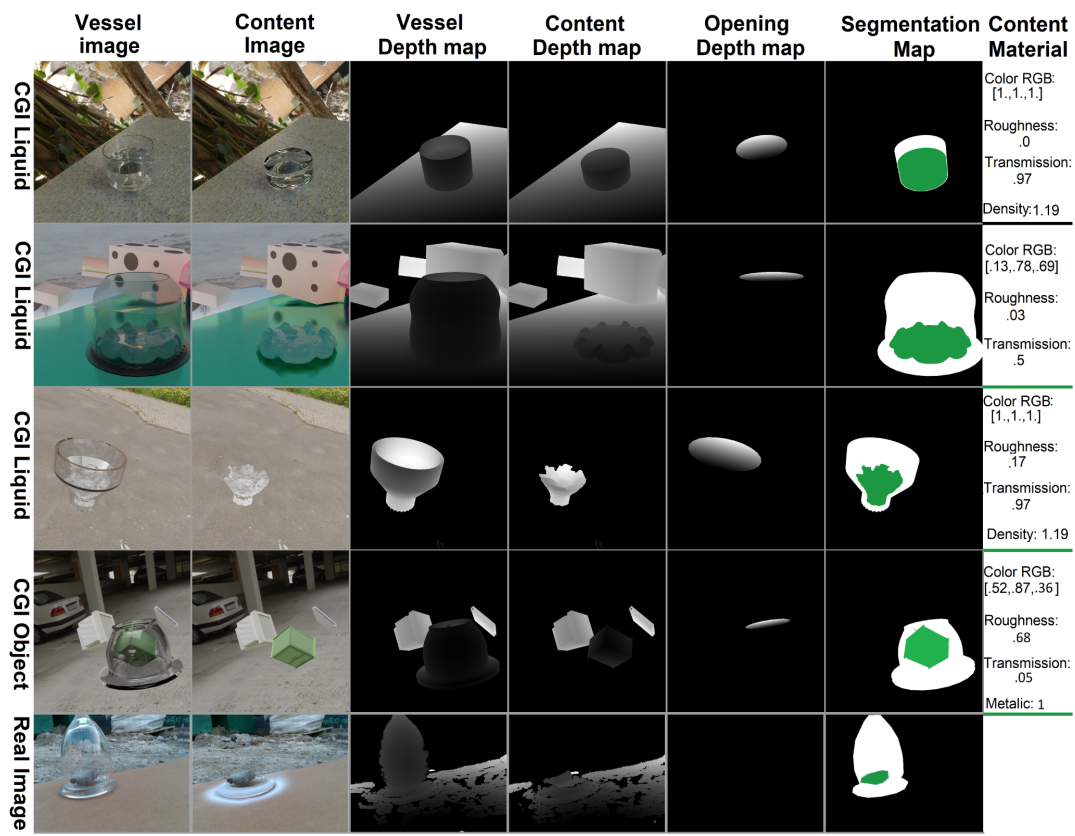

**Figure 1)** The TransProteus dataset contains images of liquids and objects in transparent containers for both simulated and real pictures. Depth maps, 3D models, 2D annotations, and properties of the materials are supplied for the vessels and the materials and objects within them.

[1]Department of Chemistry, [2]Vector Institute, [3]Department of Computer Science University of Toronto, [5]Faculty of Applied Science and Engineering [4]CIFAR Lebovic Fellow. Emails: sagieppel@gmail.com, haoping.xu@mail.utoronto.ca, yiruhelen.wang@mail.utoronto.ca, alan@aspuru.com

# 1. Introduction

Handling materials and objects inside transparent containers is essential to a wide range of applications, including chemistry labs, medicine, and material research. Visual recognition of the vessel and its content is vital for most of these applications. The lack of such understanding makes many lab tasks impossible for a robotic system and forces humans to spend considerable time on menial tasks. Transparent objects represent a significant challenge for computer vision.[1-3] Methods for 3D reconstruction and depth sensors assume that light moves in a straight line and fails on transparent objects. Machine learning approaches for computer vision tend to be assumption-free and are not affected by these problems. Deep learning-based approaches have shown promising results in extracting the 3D shapes of transparent objects from a single image.[1-3] However, training these methods demands large datasets of images annotated for the specific task. In this case, the annotation includes a 3D model of both the transparent vessel and its content (Figure 1), segmentation masks, and the properties of the materials that make up the vessel and its content (color, transparency, reflectance, roughness, etc.). This work presents a new dataset focused on these tasks, which combines computer-generated images (CGI) for training and real-world photos for testing. The dataset was generated using the Blender 3D software[4] with an emphasis on generality and diversity. Over 500 high-definition backgrounds (HDRI)[5] were used, providing a wide variety of natural illumination and environments. In addition, over 13,000 random objects[6,7] were used for both the background and the vessel content. Finally, over 1400 material textures (PBRs)[8] were used for the ground plane. The vessels were procedurally generated with an unlimited number of different curves, shapes, and materials. Two types of content were generated inside the container. The first type is random objects taken from the ShapeNet dataset[6,7] and put inside the vessel (Figure 3). The second content type was liquids with various properties simulated using the Blender MantaFlow[9] tool with effects such as splashing, foam, and bubbles (Figure 3). Container shapes and materials for both vessels and content were procedurally generated. Altogether, this makes the TransProteus dataset one of the most diverse synthetic datasets in terms of environment, illumination, materials, objects, and setting (Figure 3). In addition, we created a small dataset containing 104 real-world photos with depth maps of both the vessel and its content (Figures 1, 3); this set was created using the RealSense depth sensor[10] and used to test the net trained on the synthetic CGI dataset (Figure 3). We also introduce a new model and training method for predicting a 3D model from a single image as an XYZ map. The prediction is independent of the camera type and image source. Previous work has already addressed scale-invariant[11-13] and unknown camera parameters[14-16] when predicting depth maps from images. We expand upon this work by predicting the XYZ map instead of the depth map.[17] Hence each pixel in the prediction map contains the X,Y,Z coordinates of a point instead of the distance to this point (Figure 2). This XYZ map is equivalent to the point cloud and does not depend on camera parameters. A major issue with predicting the model as a XYZ map is that the coordinates depend on the origin point, which cannot be deduced from the image. Making the XYZ prediction independent of origin (translation invariance) is achieved by using the distance between every two points in the model as the loss metric instead of the absolute XYZ coordinates (Figure 4). The loss is simply the sum of the absolute difference (L1) between the normalized predicted distance and the Ground Truth (GT) distance between the same pixel pairs (Figure 4). The net used for this task is a simple, fully convolutional neural net (FCN) that outputs the XYZ map as an image with three values per pixel (Figure 2). Another method demonstrated here is a net that receives the image and the region of the vessel in the image and predicts the properties of the materials inside the container and the vessel's material properties (Figure 5b).

In summary, the main contributions of this work are the following:

1) The first method and dataset for predicting 3D shapes of materials, liquids, and objects inside transparent vessels.

2) A dataset and method to predict the properties of materials for both transparent containers and the things inside them. These properties include color, roughness, transparency, reflectance, and many other visual properties.

4) A novel method to predict a 3D model directly from an image as an XYZ map. The prediction is independent of the image's source and camera type.

5) A demonstration of how environments and materials repositories created for the CGI artist community can significantly increase the diversity of computer-generated synthetic datasets.

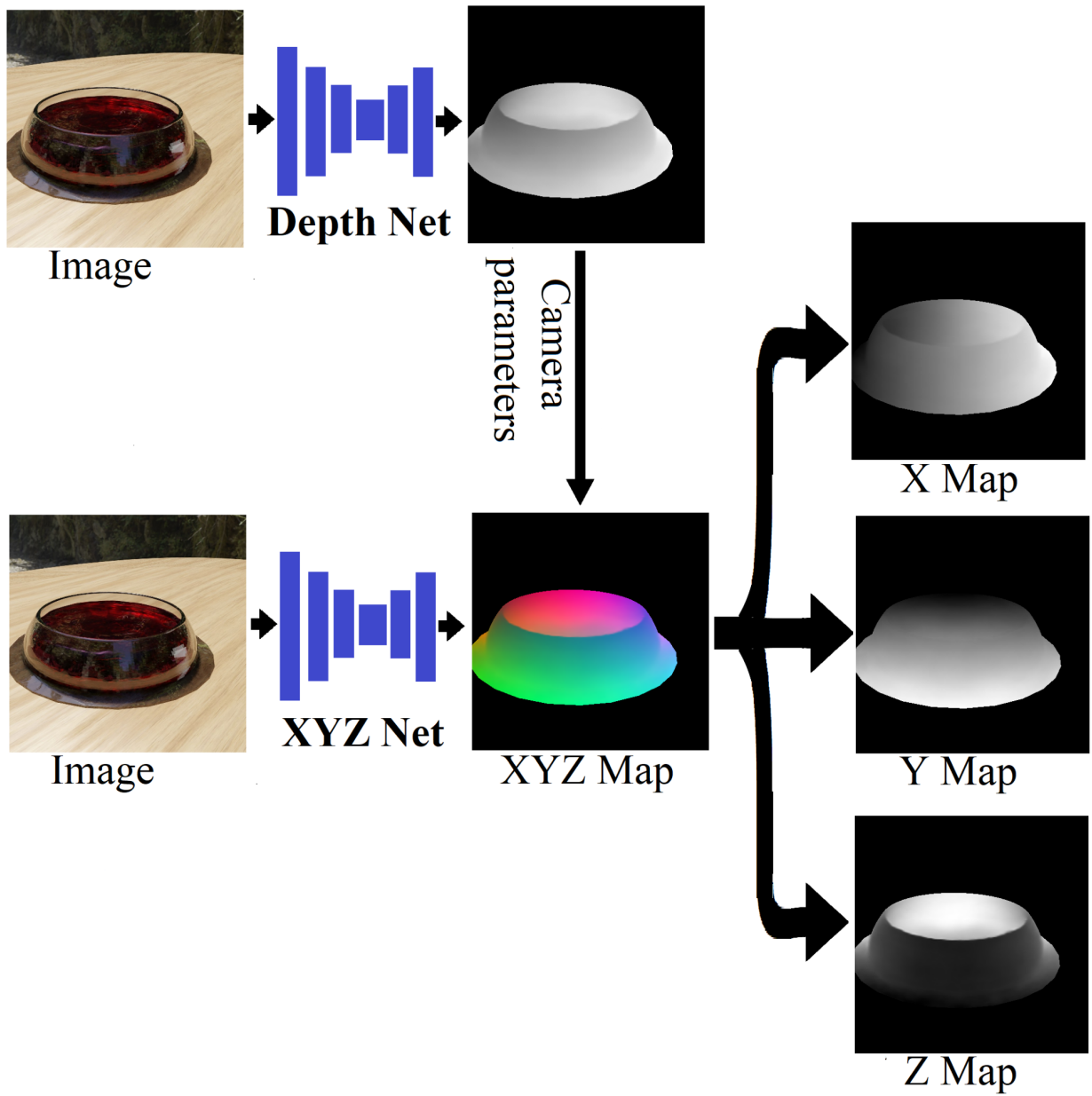

Figure 2) Depth prediction net versus XYZ prediction net. Depth maps give the distance of every pixel from the camera. Distance is encoded as the pixel intensity/brightness. The XYZ net predicts for each pixel the XYZ coordinates as three values (three-layer map). Converting a depth map to an XYZ map is only possible using known camera parameters. The XYZ map is equivalent to a 3D model and does not require camera parameters. The XYZ map is displayed as a BGR image with the blue, green, and red values of a pixel corresponding to the pixel's X, Y, and Z coordinates, respectively. It can be split into three 2D maps for X, Y, and Z coordinates (where the pixel intensity corresponds to the coordinate value along this axis).

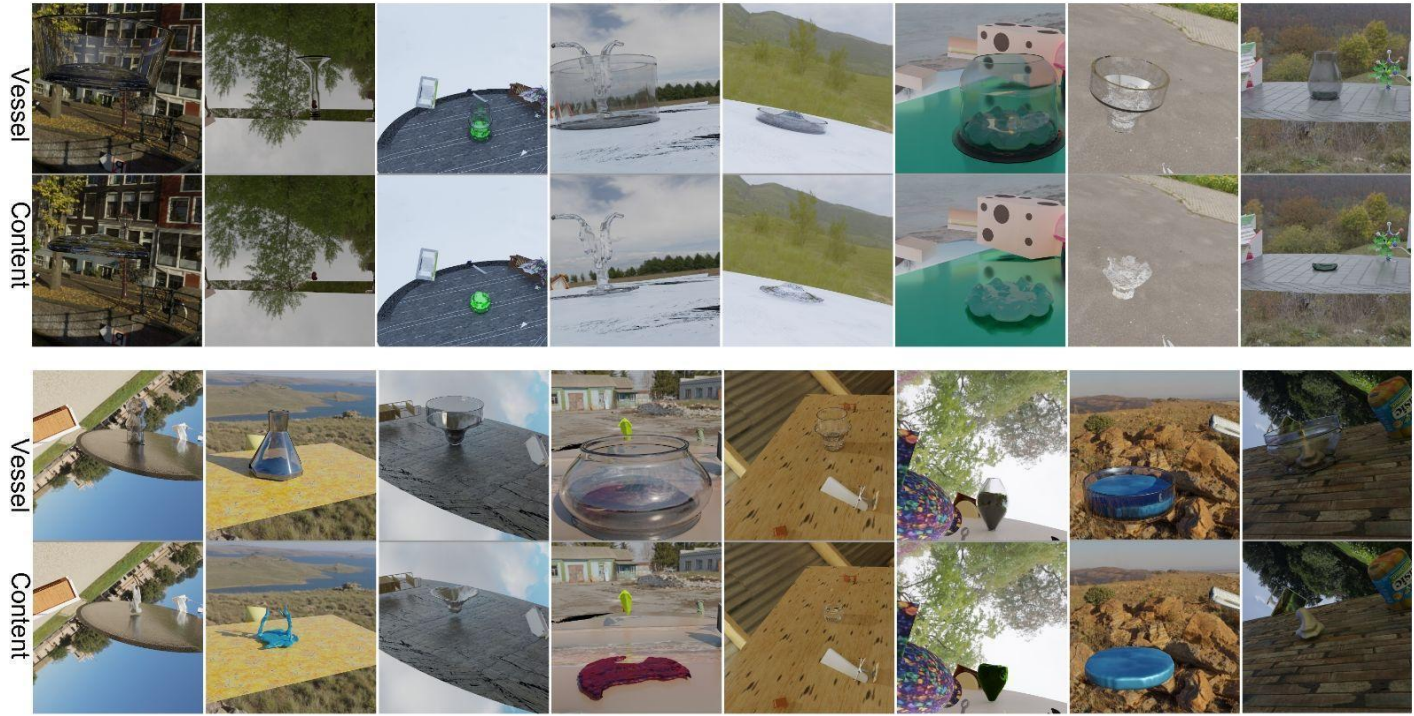

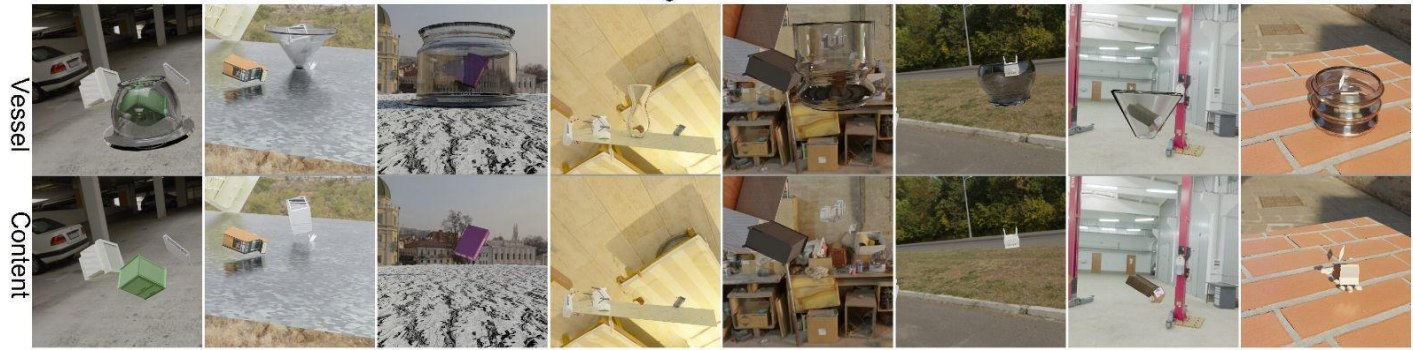

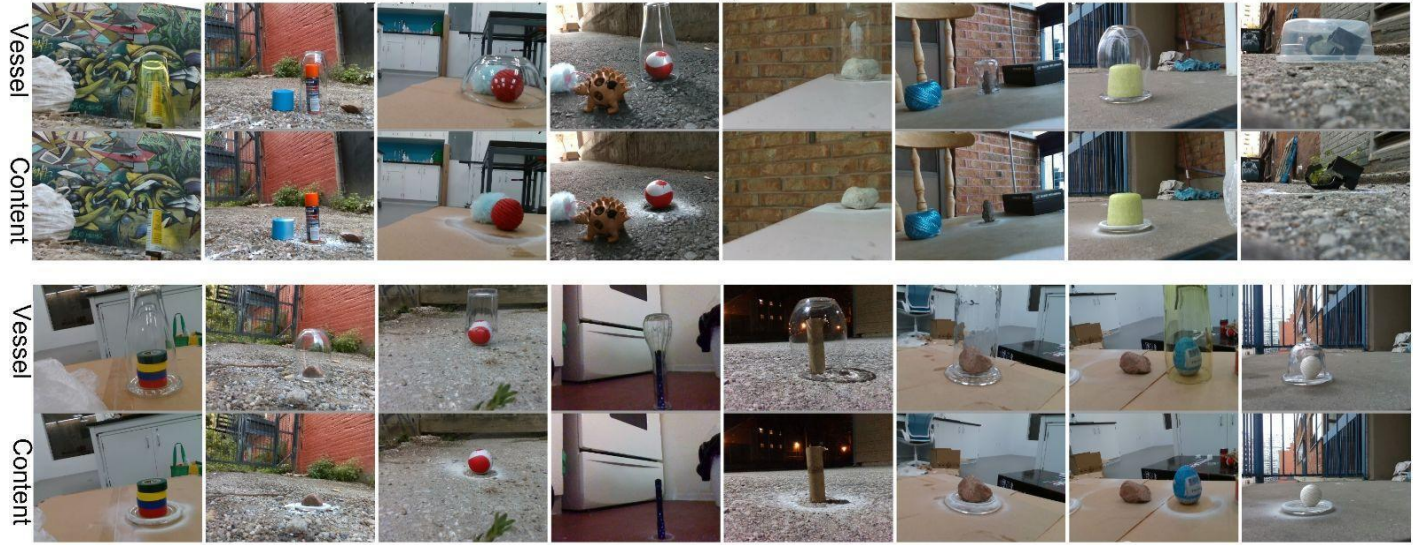

**Figure 3)** Example of simulated images for liquids and objects in vessels and real-world photos of things in containers. Each line contains the same scene, with the vessel and without the vessel (exposed content).

## 2. Related work

**Computer vision for transparent vessel content and chemistry:** Machine vision has been used for decades to recognize simple properties like color, turbidity, and fill level for materials in vessels for analytical chemistry and bottle filling.[18-27,54-56] More advanced algorithms based on methods such as graph-cut were used to segment material with unpredictable surface shapes (like solids) and multiphase materials.[28,54] However, these approaches are still limited to simple conditions with controlled environments and often fail in complex real-world scenarios. Recently, methods based on deep neural nets and convolutional neural nets (CNNs) have proven significantly more effective in performing all of the above tasks.[29-33] Semantic and instance segmentation allows one to find the region and class of each object and material phase in general conditions. However, these nets have been so far limited to 2D segmentation.[31-33]

**Predicting depth and 3D model from a single image:** Extracting a 3D model from photos could be achieved using a single image or multiple images covering different viewpoints.[11-16,34-36] The standard depth and 3D prediction methods assume that light moves in straight lines,[10,37] an assumption that fails with transparent objects.[1-3] Deep neural nets learn directly from data and therefore do not rely on any assumptions. Neural nets for extracting 3D models from images have mostly relied on using fully convolutional nets (FCNs) that predict depth maps.[11-16] The value of each pixel in the depth map corresponds to the distance of this pixel from the camera (Figure 2). This approach is assumption-free and can easily predict depth maps of transparent vessels from a single image. With known camera parameters, it is possible to convert the depth map into an XYZ map with the 3D position of each pixel in the world (Figure 2). However, without the parameters of the camera used to take the image, it's not possible to convert the depth map into a 3D model. To solve this, several methods for extracting camera parameters from unfamiliar images were suggested.[14-16] However, this usually demands additional steps. Directly predicting the XYZ map (3D coordinate per pixel) can be achieved by the same methods used for the depth map but does not demand camera parameters for creating the 3D model (Figure 2). Predicting a 3D model from a single image, as an XYZ map, was done in previous work.[17] However, in this case, the loss function was based on converting the XYZ back to depth. Since this conversion depends on camera parameters, this led to the loss of the camera agnostic property of the net.

**CGI and real photo dataset creation:** Deep learning approaches for computer vision are strongly reliant on training data. Generating data for specific tasks remains the main challenge in applying computer vision to new fields and improving the performance of existing fields. Creating a dataset can be done manually by collecting images and using humans for annotation. This approach mainly applies to classification[38] and segmentation[29-33] but can also be used for depth estimation by asking people to estimate the relative distance to two objects.[13] Other methods rely on metadata or sensor data from depth sensors, LIDAR, stereo, or a structure from motion.[34-36] These approaches fail on transparent objects.[1-3] Synthetic datasets use simulation and computer-generated imagery (CGI) to create the training data.[3,39-42] The advantage of this approach is that it is not limited by sensors and human perception and can work in any case where the data can be simulated. However, training using this dataset often gives inferior results compared to training from real data, mainly because the simulation often misses many of the complex visual features of the real world. Such datasets have been created for autonomous driving, transparent objects, liquid dynamics, and material properties.[3,39-42] However, as far as we know, no such dataset was suggested for materials inside transparent containers. Another issue with existing synthetic datasets is the use of a small, limited set of objects and environments. This makes nets trained on these datasets very limited in terms of the domain in which they can be used. Recently projects like Poly-Haven[5], ambientCCG[8], and ShapeNet[6,7] created huge repositories for objects, environments, and material textures for the CGI artist community. Using these can dramatically increase the diversity of synthetic datasets.

**Transparent object datasets:** Datasets for the segmentation of transparent objects in real-world images have been mostly created by manual annotation and image matting.[43-46,57] The largest of these datasets is Trans10k,[45] with 10k images in which the region of the transparent object is marked. The LabPics dataset contains 8k images of mostly transparent vessels in labs, hospitals, and other settings.[31,32] The vessel's content and transparent regions are manually annotated. For 3D and depth maps of transparent objects, there is still a limited number of datasets. The ClearGrasp dataset contains mostly simulated 3D data for transparent objects but with no content.[3] This dataset also collected 3D scans of real transparent objects sprayed with opaque spray and then scanned with a RealSense depth sensor. Both approaches are used in this work as well.

# 3. Dataset Generation

The goal of the TransProteus dataset is to allow the prediction of the 3D shapes and properties of materials, liquids, and objects inside vessels regardless of the application. However, the large number of research fields, industrial applications, and everyday life activities for which this problem is relevant means that it's impossible to simulate all the different objects and materials that can occur within the vessel, even for a specific field like experimental chemistry. To address this, we try to make the dataset as general as possible, assuming that if the dataset contains diverse enough examples, any network trained on this dataset will be able to generalize to new and unfamiliar systems. This means that instead of narrowing down the generated data to be as realistic as possible for a given use case, we try to make it as diverse as possible, even if many of the examples are unlikely to exist in reality. For example, the environments for both the real and simulated datasets include a large variety of backgrounds, including fields, parking lots, and many other locations that are not usually used to handle transparent containers (Figure 3). In addition, the content of each vessel was chosen to be a random object or liquid with random properties. Many of these examples are very unlikely to appear in reality (Figure 3). However, the variability and diversity of the dataset mean that any network that will learn to predict the shape and properties of all of these different cases will have to be highly generalized and work in almost any case. Code for generating the dataset is supplied.

## 3.1. Computer-generated environments

The simulated dataset was generated using Blender 2.93. Creating background and illumination for the scenes was done using high dynamic range images (HDRI). These high-definition images completely surround the scene, providing background and full natural illumination from all directions. Five hundred HDRIs were downloaded from the Poly Haven project. The HDRI environments include indoor and outdoor settings in cities, nature, and other environments (Figure 3). To further increase diversity, the HDRIs were rotated, and their intensity increased or decreased randomly for every image. To make the scene even more diverse, we use the objects from the ShapeNet dataset. We used 13k different objects from a large number of categories. Up to 10 random objects were randomly scattered in every scene, with random scale position and rotation. A ground plane was generated by creating a simple plane below the objects and assigning random physically based rendering (PBR)[47] materials to this plane.[3] These PBR materials contain realistic complex textures, displacements, and other properties of real-world materials. About 1416 different material textures were downloaded from the ambientCG project. We note that the steps taken so far are not specific to this dataset and can be used to create a general setting and environment in any synthetic dataset. They can greatly benefit synthetic datasets that tend to use a limited set of environments and settings. Generating the synthetic data was done using scripts that create random procedurally generated scenes that are different in every image. Code for generating the dataset is supplied here.

## 3.2. Procedurally generating vessels

Glassware shapes vary widely between different use cases. However, almost all glassware tends to have cylindrical or symmetric shapes (from a top view). Hence, we can describe every vessel's top view as a circle or other symmetric shape, and the vessel curvature (profile) is some 2D function (Figure 3). The curvature (profile) was generated by randomly combining linear, polynomial, and sinusoidal functions to create the vessel curvature derivative, leading to a random but mostly smooth 2D function. This leads to a wide range of shapes that seem to cover any vessel we encounter in the real world and many that we do not (Figure 3).

## 3.3. Generating liquids and objects for vessel content

The contents of the vessel were generated using one of three methods. The first approach was to take a set of random objects from the ShapeNet dataset and randomly position them inside the vessel. This is by far the most diverse method for filling the vessel but the least realistic (Figure 3, Center). The second approach was to create a random blob of liquid inside the vessel. The liquid was given a random shape, properties, and initial velocity and was simulated using the MantaFlow module of Blender. Images were captured in various steps of the simulation. This creates a wide variety of shapes associated with liquid splashing, spilling, and sticking, which covers the wide range of liquid behaviors in the real world (Figure 3, Top). The MantaFlow liquid simulation also contains tools for simulating foam, and bubbles, which were used. The final approach for content creation is to represent the static liquid as a mesh with a flat surface that fills the bottom part of the vessel (Figure 1, Top). No actual liquid simulation is needed in this case. This is the most common way liquid will appear in vessels, but it has the least diversity.

## 3.4. Generating and assigning materials

Materials for both the vessel and its contents were generated using the Principled BSDF material[48] shader in Blender. This tool enables the control of all the visual properties of the material, including color, transmission (transparency), metallic (reflection), IOR, roughness, luminescence, and many others. The materials generated by this tool are uniform, and unlike PBR materials, they do not have complex textures (Figure 3). However, glass vessels and liquids tend to be very homogenous anyway. The advantage of BSDF materials over PBR materials is that all the visual properties are given as a list of numbers with constant length.[48] This can be saved and later predicted by the net. For the case of objects inside the vessels, the objects are already supplied with materials and textures (from the ShapeNet dataset). Therefore, the materials of objects inside the vessels were kept as they are for 50% of cases and replaced by uniform Principled BSDF materials in the remaining 50%. For liquids and the vessel, randomly generated BSDF materials were used for all cases. It should be noted that liquid splashing, foam, and bubbles effects also influence the liquid material textures. These properties are not described by the BSDF shader and cannot be predicted using the dataset.

## 3.5. Creating a real-world image dataset

Generating real-world images and 3D scans of transparent objects and their content is challenging for two main reasons. First, standard depth sensors like LIDAR and structure light (RealSense[10] and Kinect) do not work on transparent vessels. Second, scanning the depth map of the vessel content requires a method to remove the vessel without moving its content or the sensor (because depth sensors can't penetrate the vessel surface). The problem of 3D scanning transparent vessels was solved by first taking an image of the vessel and then spraying it with opaque spray without moving the vessel or the sensor, similar to the method used in the ClearGrasp dataset.[3] Removing the vessel without moving or changing its content is a more challenging task. To achieve this, the vessel was put upside down over an object, scanned, and then removed without moving the object or sensor (Figure 3, Bottom). This approach is clearly valid only for solid content. In addition, it relies on the ability of the nets to work with an arbitrary orientation of the vessel. The images were taken in various buildings, parking lots, and yards, with a large set of random vessels and objects (14 locations, 20 vessel types, and 25 different objects inside the vessels). The real sense D435 depth sensor was used for scanning, and AESUB blue spray was used for painting the vessels. This procedure is time-consuming. Therefore, only 104 images were collected and used for testing the dataset. The annotation of the vessel and content masks was done manually on the images of the vessel and exposed content. The RealSense[10] depth data is very noisy. The depth map was cleaned by removing points that are more than 10 cm from the object center. Since all the vessels used are smaller than 10 cm, a distance of more than 10 cm from the center implies error in measurement.

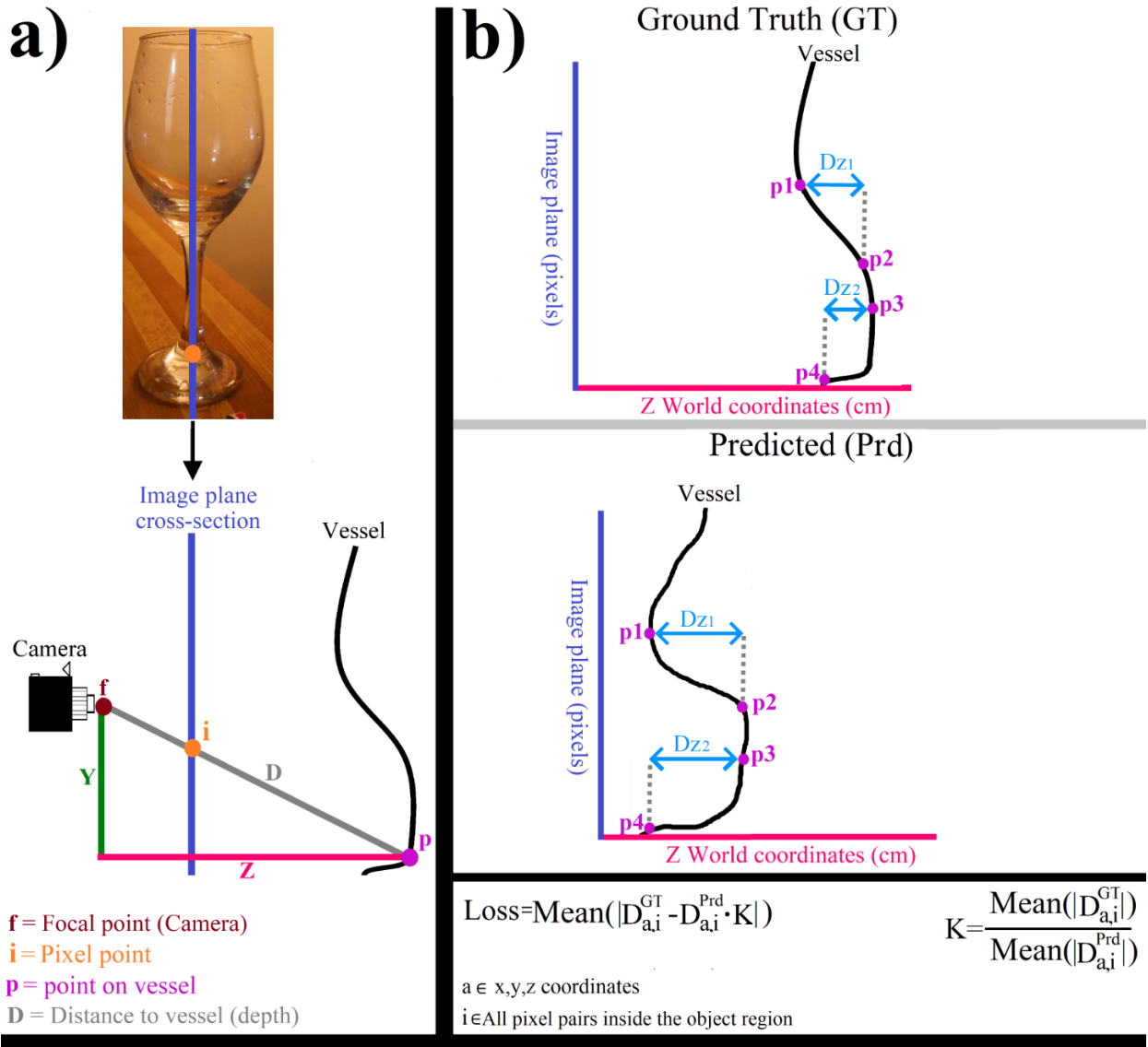
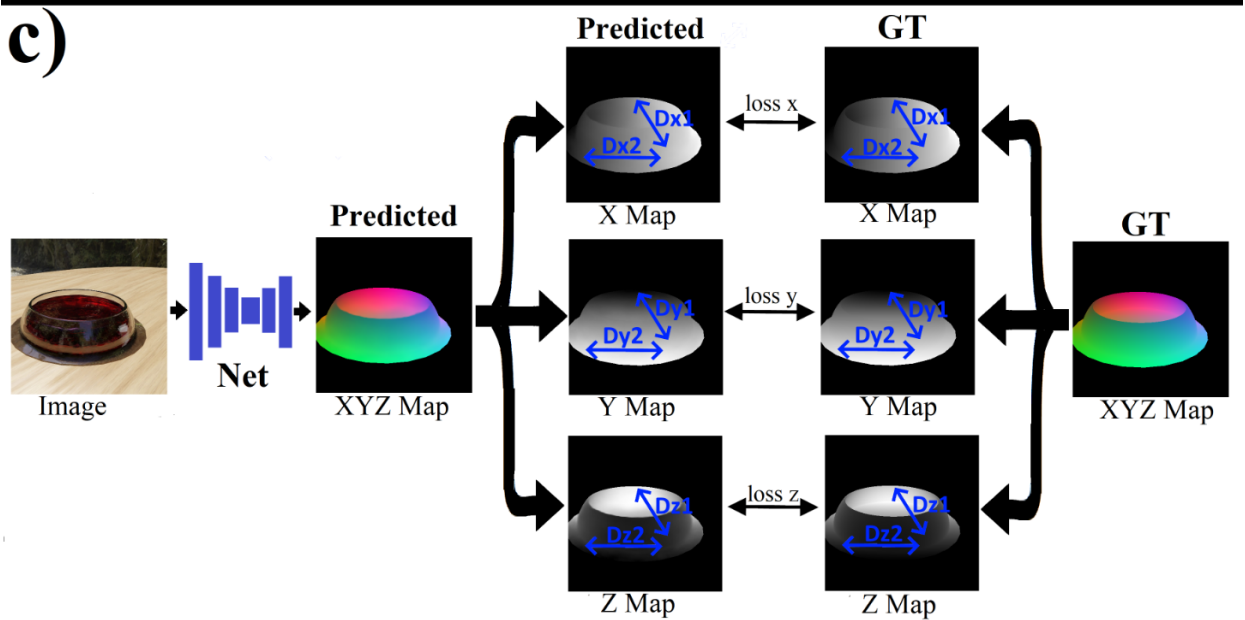

**Figure 4)** Predicting 3D model as XYZ map and difference-based loss function. a) Cross-section of the predicted map along one image column (blue line). b) Profiles of the predicted and GT maps in the Z coordinates. The predicted and GT maps' translation and scale are inconsistent with each other. Therefore, the distances between the points (Dx, Dy, Dz) are used for the loss function instead of the absolute x,y,z coordinates of the points (making the translation irrelevant). c) Loss display on the 2D XYZ images. The distances (Dz, Dy, Dx) refer to the difference between the two points in the x, y, and z coordinates, respectively (and NOT to the distance in pixels between the two points on the image plane).

# 4. Predicting 3D model as an XYZ map

Predicting the 3D model as an XYZ map can easily be done using FCN that receives an image and predicts the XYZ map as a three layer matrix (Figure 2).

## 4.1. Translation invariance loss

Since the coordinates of the XYZ points have an arbitrary origin point (assuming camera parameters are unknown), it is necessary that the loss will be independent of the origin point (translation invariant). Translation invariant loss can be achieved by using the distance between points in the model as the metric instead of the point's X,Y,Z coordinates (Figure 4b). This is because the distances between two points in the model do not depend on their absolute coordinates or the origin point.

This is illustrated in Figure 4b: assume that the vertical axis is some line in the image plane (in pixels) and the horizontal axis is the Z coordinates in cm. The predicted map is translated relative to the GT map, leading to completely different Z coordinate values for GT and predicted maps, even when the shapes are similar (Figure 4b). However, the difference in Z coordinates ($D_{z,1}$) between two points (p1, p2) is independent of origin. Therefore similar GT and predicted shapes should have the same $Dz$ value (Figure 2b).

If $D_{z,i}^{GT} = Z_1^{GT} - Z_2^{GT}$ is the difference between the Z coordinates in pixels 1 and 2 in the GT map (Figure 4b), and $D_{z,i}^{prd} = Z_1^{prd} - Z_2^{prd}$ is the Z difference between the same two pixels in the predicted XYZ map, then $|D_{z,i}^{prd} - D_{z,i}^{GT}|$ is the translation independent error/loss.

The loss function is, therefore the absolute mean of differences between these distances, along each of the axes and for every pair of pixels in the object:

$$Loss = Mean(|D_{a,i}^{GT} - D_{a,i}^{Pr}|)$$

$i \in$ all pairs of pixels inside the objects
$a \in X, Y, Z$ axes.

## 4.2. Scale-invariant loss

The above loss is translation invariant but still scale-dependent. This means that if the predicted model has the same shape but a different scale than the GT model, the error will be high. In our case, the scale of the 3D model cannot be extracted from the image. We, therefore, want the loss to be independent of the predicted model scale. This can be solved by adding a scale normalization constant to the predicted distance (Figure 4). This basically means finding the scale difference between the predicted and GT models and rescaling the predicted model to match the GT model's scale.

Finding the scale factor ($K$) could be done by taking the ratio between the sum absolute distances ($D$) between every pair of points (and along each axis) in the GT and predicted XYZ maps:

$$K = \frac{Mean(|D_{a,i}^{GT}|)}{Mean(|D_{a,i}^{Pr}|)}.$$

$i \in$ all pairs of pixels inside the objects
$a \in X, Y, Z$ axes.

This leads to a scale-invariant loss function:

$$Loss = Mean(|D_{a,i}^{GT} - K \cdot D_{a,i}^{Pr}|).$$

Note that this was done only to positive distance ratios $\frac{D_{a,i}^{GT}}{D_{a,i}^{Pr}} > 0$. The scale factor (K) is calculated once for the entire image and multiplied by the predicted distances to match them to the GT distances. In addition, we want to avoid relative scales (K) which are too big or small. This is because very large or small relative scales cause the training to explode or get stuck.

We, therefore, add a scale controlling term to the loss function that is used only if the scale factor (K) is larger than ten or smaller than 0.1. If $K > 10$, this term is $Mean(K)$, which causes the scale ratio to decrease, while if $K < 0.1$, we add the term $-Mean(K)$ to the loss function, which forces the prediction scale (K) to increase. This extra loss element guarantees that the scale factor (K) will always be in the range of 0.1–10.

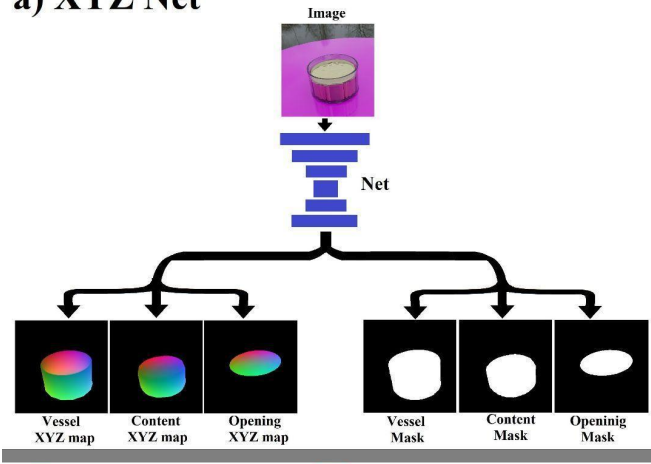
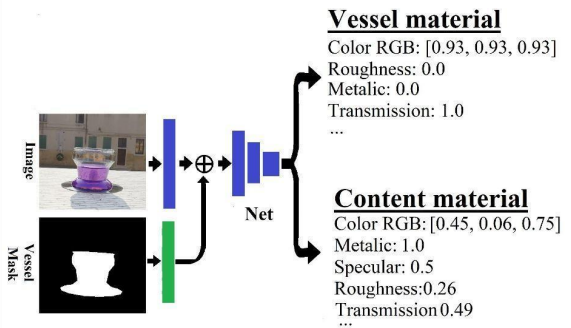

**Figure 5)** Structure of neural nets. a) XYZ prediction net is a simple FCN that produces the XYZ map as a three-layer map with X,Y,Z values per pixel. The objects' masks are predicted as two-channel probability maps with two values per pixel (belongs/does not belong to the object). b) The material property prediction net is a simple convolutional net (Resnext) that receives the image and the vessel mask (region) and predicts the properties of the vessel and content materials as a vector.

# 5. Predicting material properties

The material properties are given as a list of numbers that include RGB color, transmission (transparency), roughness, metallic (reflectiveness), IOR, and others (Figures 5b). Standard convolutional neural nets for image classification can easily be modified to predict these values by using the final output vector of the net to represent these properties (Figure 5b). This was done for the material properties of both the vessel and its content. The vessel region was added as an input for the net by processing it using a single convolutional layer and adding the result to the first layer of the convolutional net (Figure 5b). The training was done using standard ResNet[49] training methods. The loss function was the sum of the absolute difference between the predicted and GT vectors.

# 6. Training with additional datasets

To improve results on real-world images, we also use vessel mask and content regions from the LabPics dataset as additional training data for vessel and content mask prediction. These were applied in 30% of the training steps (The XYZ map loss was set to zero in this case).

# 7. Evaluation

## 7.1. 3D model evaluation

Evaluation of the XYZ map prediction was done by modifying three standard metrics. The mean absolute error (MAE)[11] takes the mean Euclidean distance between the predicted and GT points for every pixel belonging to the object:

$$MAE = Mean(D(p_i^{GT}, p_i^{Prd}))$$

Where D is the Euclidean distance between the points $p_i^{GT}, p_i^{Prd}$ for the same pixel (*i*) on the GT and predicted XYZ maps and $i \in$ all the pixels in the object region.

In addition, since scales are arbitrary, the MAE value has little meaning. We, therefore, normalized the MAE by dividing it by one of two values: The mean absolute deviation (MAD) is the mean distance between points in the GT object and the GT object's center, and it was used as the first normalization factor:

$$MAD = Mean(D(p_i^{GT}, c^{GT}))$$

Where, $c^{GT}$ is the center (average) of the GT points.

In addition, we use the maximum distance between two points in the GT object (*MaxDst*) as a second normalization factor for the MAE:

$$MaxDst = Max(D(p_i^{GT}, p_j^{GT}))$$

$i, j \in$ all pixels in the object region.

An additional metric is the standard R-squared, which uses the sum of the squared Euclidean distances between the predicted and GT points for every pixel divided by the sum of the mean squared distance between the GT points and the GT object's center:

$$R^2 = 1 - \frac{RSS}{TSS}, \text{ with } RSS = \sum_i^n D(p_i^{GT}, p_i^{Prd})^2, \text{ and } TSS = \sum_i^n D(p_i^{GT}, c^{GT})^2$$

An additional metric is the Chamfer distance,[53,] which is calculated by finding for each point in the GT object the closest predicted point (Euclidean distance) and finding the mean of this distance:

$$d_{cd}(S_1, S_2) = \sum_{x \in S_1} ||Min_{y \in S_2}(D(x,y))|| + \sum_{x \in S_2} ||Min_{y \in S_1}(D(x,y))||,$$

where $S_1, S_2 \subseteq R^3$ for the predicted and GT point clouds, respectively.

Note that the Chamfer distance ignores the position of the points on the image grid and uses only their XYZ position. As in the case of the MAE, the distance was normalized by both the mean deviation (MAD) and the maximum deviation (*MaxDst*) between points on the GT object.

## 7.2. Effect of scale and translation on error

The predicted XYZ map is scaled and translated to match the GT object (Section 4). This normalization removes two types of errors (scale and translation) and leaves only the errors resulting from the object shape. However, the scaling and translation factors are found using the vessel object and apply to the content object. Therefore, the content prediction still contains both scaling and translation errors. In principle, this is not a problem since we expect the scale and translation factors of the vessel and content to be the same (otherwise, the content will be out of proportion and position to the vessel containing it). However, it's also interesting to isolate the shape error of the content. To achieve this, we also calculate the scale and translation factors using the content object (instead of the vessel) and use these to scale and translate the predicted content object to match the GT content.

## 7.3. Evaluating material properties prediction

Since every property of the material is represented as one or more numbers, the prediction accuracy was evaluated using the mean absolute error (MAE) between the predicted and GT material properties. This is simply the mean absolute difference between the predicted and GT values. Note that all the predicted properties have a value ranging between 0 and 1. Hence, the MAE could also be considered as the difference between the predicted and real values by percentage. It was not possible to collect these material properties from real images; therefore, the evaluation is based on the CGI images only.

## 7.4 Evaluation of 2D instance and semantic segmentation

For evaluating the segmentation of the regions belonging to the content, vessel, and opening in the image, we choose the standard IOU metrics. The intersection over union (IOU) is the main metric used to evaluate semantic segmentation and is calculated separately for each object. The intersection is the sum of the pixels that belong to the object according to both the net prediction and the dataset ground truth (GT), while the union is the sum of pixels that belong to the object based on either the net prediction or the GT. The IOU is the intersection divided by the union. The recall is the intersection divided by the sum of all pixels belonging to the object according to the GT annotation, while precision is the intersection divided by the sum of all pixels belonging to the object based on the net prediction.

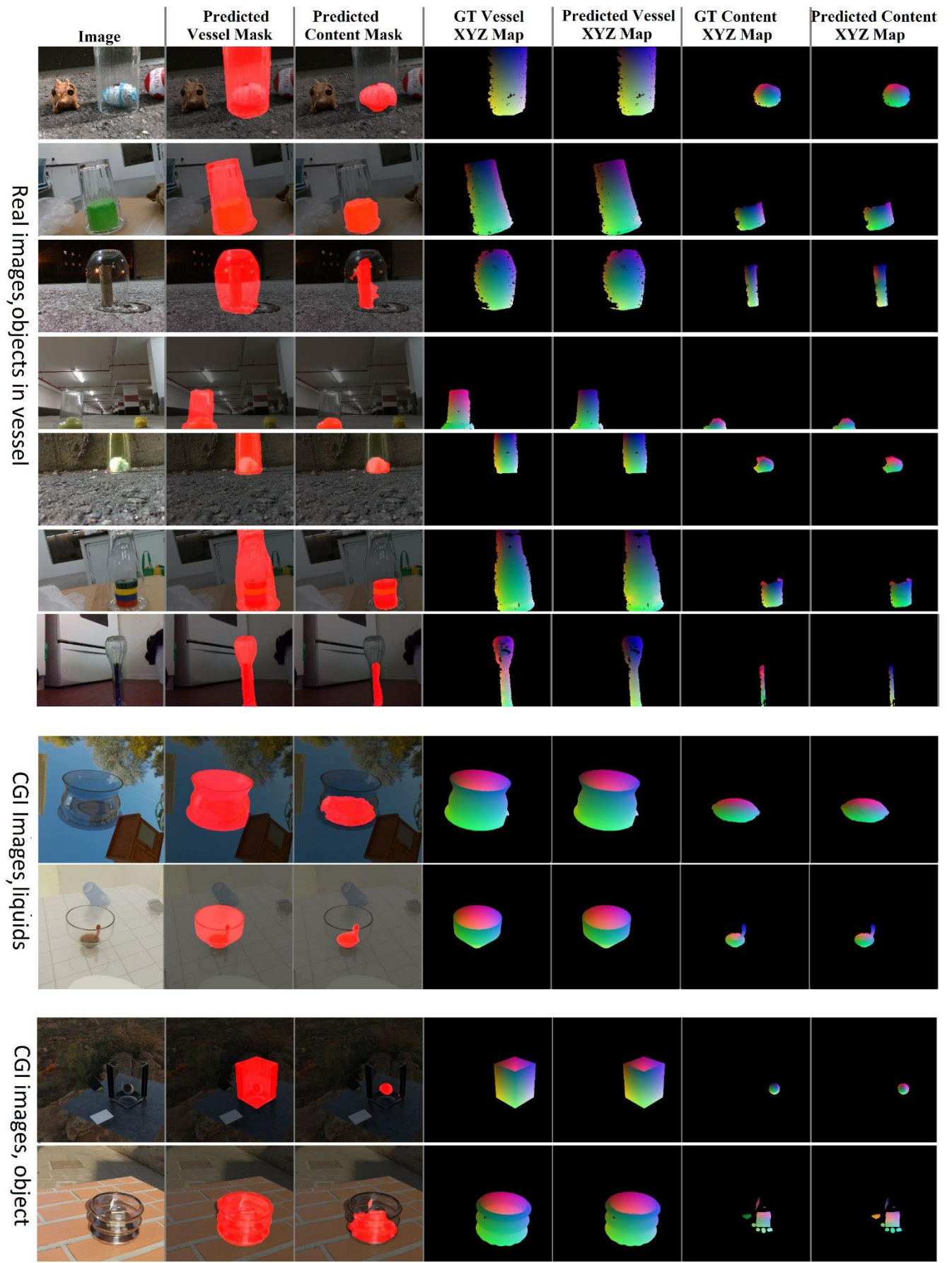

Figure 6) Results of the net prediction for XYZ map and Segmentation map for simulated and real images.

# 8. Results

## 8.1. Results for 3D model prediction

The results for the 3D model XYZ prediction appear in Tables 1 and 2 and Figure 6. It can be seen from Table 1 that the net achieves good accuracy for predicting the 3D shape of the vessel for both real and simulated pictures. This is true even when the net is trained only on CGI images and tested on real photos (Table 1). Training with additional real images (LabPics dataset) gives only a minor advantage in this case. For content prediction in real-world images (Table 2), the net gives good accuracy for predicting the 3D shape of the object in the vessel (Section 7.2). However, when the scale and translation are included in the error, the accuracy of the prediction is significantly lower (Table 2). This can be explained by the fact that the exact position and scale of the object in the vessel are often hard to determine from the image (Figures 3, 6), while the object's shape is usually clear. The net achieves good accuracy in predicting the shape of liquids inside the vessel even when including translation and scale errors (Table 2, Figure 6). This can be explained by the fact that liquids tend to either completely fill the bottom portion of the vessel or stick to the vessel surface (Figure 3). In both cases, the position (translation) inside the vessel is clear. For simulated objects inside vessels, the net achieves medium accuracy for shape prediction (Table 3, Figure 6). However, adding translation and scale errors did not significantly affect the accuracy (Table 3). This can be attributed to the significant variance in the shape of the object used (Figure 3). The vessel opening plane was predicted with high accuracy (Figure 1, Table 1), similar to the vessel shape predictions. This makes sense, given that the vessel opening can be viewed as the top part of the vessel.

## 8.2. Results for material properties prediction

The results of material properties prediction are given in Table 3. It can be seen that the net achieves good accuracy with a mean absolute error of less than 10% for all properties. The vessel material is also predicted with high accuracy (Table 3), but in that case, the range of the material properties of the vessel surface is narrow, making the prediction relatively easy. For the content material properties, the variance in the generated materials is high in all properties (color, transparency/transmission, roughness, and metallic/reflectiveness). Also, the difference in illumination and vessel surface reflection is quite significant (Figure 3). Even so, the net achieves good accuracy for all properties, implying that it learns to compensate for background illumination and the vessel surface.

## 8.3. Segmentation results

The results for the segmentation of vessels and content are given in Table 4 and Figure 6. The net predicts the vessel region with high accuracy (IOU > 80%) and the content region with medium accuracy (IOU > 50%) for real and simulated images (Figure 6). Training the net using a combination of the virtual TransProteus images and the real images of the LabPics dataset gave a 7% improvement for the vessel 2D segmentation in the real photos but only a 3% improvement for the vessel content for the same photos (Table 4). It should be noted that the task of the 2D segmentation of transparent containers and their content is also covered by the LabPics dataset. However, the LabPics dataset predicts the shape of the content as it is viewed through the vessel's transparent surface (distorted shape). On the other hand, the TransProteus dataset predicts the undistorted content shape, which is the region of the object as it would be viewed if the vessel was not in the way (Figure 1).

# 9. Conclusion

This work demonstrates the first method and dataset for predicting the 3D shape and properties of materials, liquids, and objects inside transparent containers. The nets achieve good results for simulated and real-world images, but considerable challenges remain in terms of prediction accuracy and evaluation methods. We also introduce a simple method to predict a 3D model from an image using a neural net that is independent of camera parameters and can work with images from unknown sources. For the creation of the dataset, we use existing methods for rendering and simulation. However, we combine this with large textures repositories and HDRI repositories used by CGI artists. These repositories are relatively unutilized in the machine learning community, and using them dramatically increases the diversity and generality of the dataset, making it one of the most general synthetic datasets in terms of environments, materials, and objects. As a result, the net trained on this synthetic data alone achieves good results on 3D and 2D shape prediction for real-world complex images. Future challenges include expanding the dataset for multiphase systems (phase separating liquids, suspension), simulating more complex chemical systems, and increasing the prediction accuracy.

**Table 1: Results for XYZ net on Vessel 3D model prediction (Section 7.1)**

| Test Data | Net/Training | MAE MDV | MAE MaxDist | Chamfer MAD | Chamfer MaxDist | $R^2$ |
|---|---|---|---|---|---|---|
| Real images Object vessels (RealSense) | XYZ Net (with LabPics) | 12.1% | 2.5% | 15.9% | 3.3% | 0.96 |
| | XYZ Net (Only TransProteus) | 14.6% | 2.9% | 17.4% | 3.5% | 0.94 |
| Simulated Liquid in a vessel | XYZ Net (with LabPics) | 7.3% | 1.2% | 7.6% | 1.2% | 0.97 |
| Simulated object in a vessel | XYZ Net (with LabPics) | 8.0% | 1.7% | 9.5% | 1.7% | 0.97 |
| Vessel Opening Plane | XYZ Net (with LabPics) | 11.7% | 1.5% | 9.8% | 1.2% | 0.95 |

**MAE:** Mean absolute Euclidean distance between GT and predicted points on the same pixel pairs.
**MAD:** Mean absolute deviation of the GT. Mean Euclidean distance between points on the GT object and the GT object center (X,Y,Z average).
**MaxDist:** Maximal Euclidean distance between any two points on the GT object.
**Chamfer distance:** Mean Euclidean distance between each point on the GT object and closest predicted point, plus mean Euclidean distance between each point on the predicted object and the closest GT point.

**Table 2:** Results for XYZ net on Content 3D model prediction (Section 7.1) for predicted content scale/translation normalized by vessel object and normalized to match GT content object.

| Test Data | Net/Training | Content XYZ (Normalized to Content scale) | | | | | Content (Normalized to vessel scale) | | | | |
|---|---|---|---|---|---|---|---|---|---|---|---|
| | | MAE MDV | MAE MaxDist | Chamfer MAD | Chamfer MaxDist | $R^2$ | MAE MAD | MAE MaxDist | Chamfer MAD | Chamfer MaxDist | $R^2$ |
| Real images Object vessels (RealSense) | XYZ Net (with LabPics) | 21.9% | 2.8% | 30.1% | 3.9% | 0.89 | 52.7% | 2.9% | 74.8% | 6.5% | 0.27 |
| Real images Object vessels (RealSense) | XYZ Net (Only TransProteus) | 22.8% | 3.0% | 30.8% | 4.0% | 0.87 | 54.0% | 6.6% | 76.0% | 9.4% | 0.36 |
| Simulated Liquid in a vessel | XYZ Net (with LabPics) | 18.7% | 2.3% | 17.5% | 2.0% | 0.82 | 21.8% | 2.5% | 20.0% | 2.2% | 0.78 |
| Simulated object in a vessel | XYZ Net (with LabPics) | 36.0% | 3.6% | 36.0% | 3.6% | 0.73 | 38.5% | 3.6% | 45.2% | 4.3% | 0.62 |

**MAE:** Mean absolute Euclidean distance between GT and predicted points on the same pixel pairs.
**MAD:** Mean absolute deviation of the GT. Mean Euclidean distance between points on the GT object and the GT object center (X,Y,Z average).
**MaxDist:** Maximal Euclidean distance between any two points on the GT object.
**Chamfer distance:** Mean Euclidean distance between each point on the GT object and closest predicted point, plus mean Euclidean distance between each point on the predicted object and the closest GT point.

**Table 3:** Material properties prediction mean absolute error (MAE) for content and vessel surface (Section 7.3).

| | Mean Absolute Error (MAE) | | |
|---|---|---|---|
| **Property** | **Content Liquid** | **Content Object** | **Vessel Surface** |
| **Transmission/Transparency** | 2.1% | 5.6% | 0.5% |
| **Color (RGB)** | 10.7% | 4.1% | 4.6% |
| **Metallic/Reflectiveness** | 2.9% | 4.8% | 0.7% |
| **Roughness** | 5.3% | 3.9% | 0.6% |

**Table 4:** Results of semantic segmentation on real images and simulated images (Section 7.4) for net training on only TransProteus, and for a net trained on TransProteus combined with the LabPics dataset (real photos).

| Evaluated on | Net/Training | Vessel Segmentation | | | Content Segmentation | | |
|---|---|---|---|---|---|---|---|
| | | mIoU | Precision | Recall | mIoU | Precision | Recall |
| Real images Object vessels (RealSense) | XYZ InNet (with LabPics) | 87% | 90% | 96% | 55% | 70% | 71% |
| Real images Object vessels (RealSense) | XYZ Net (Only TransProteus) | 80% | 90% | 87% | 52% | 68% | 69% |
| Simulated Liquid in a vessel | XYZ Net (with LabPics) | 98% | 99% | 98% | 84% | 87% | 96% |
| Simulated object in a vessel | XYZ Net (with LabPics) | 96% | 98% | 98% | 63% | 74% | 81% |
| Vessel Opening plane (CGI) | XYZ Net (with LabPics) | 94% | 96% | 98% | | | |

# 11. Appendix

## 11.1. Implementation details

The XYZ net was implemented using a standard FCN (DeepLab)[50,51] with Resnet101[49] encoder, ASPP dilated convolution decoder, and three layers of skip connection + upsampling (UNet hourglass structure[52]). The final layer of the net was split into predicting the XYZ maps of the vessel, the vessel content, and the vessel opening (Figure 5a). Each of these maps includes three layers which give the X,Y,Z coordinates for each pixel (Figure 5a). The loss for each map was calculated as described in Section 4. In addition, the regions of the vessel and content in the image were predicted as 2D masks (Figure 5a). Each mask was predicted as a two-layer probability mask (pixel belongs/does not belong to the object). The loss for each mask was calculated using the standard per pixel cross-entropy function. The net was trained on a single RTX 3090 GPU. The training and net structure were the same as for a standard FCN for semantic segmentation. The PyTorch implementation and trained models have been [made available](). Image augmentation included resizing, cropping, blurring, decoloring, and adding white noise but NOT mirror reflecting and rotation. The net was trained once for all content types, with 40% of the training steps using simulated objects as the vessel content, 40% using simulated liquids as the vessel content, and the remaining 20% using liquids with a flat surface (Section 3.3).

## 11.2. Hierarchical loss for vessel and content 3D shape

We want to predict the XYZ map for the vessel and its content as well as the vessel opening surface (Figure 5a). They can be considered as 3 different overlapping objects Predicting XYZ maps for these three objects independently using the loss function in Section 4.2 will lead to different scales and translations for each object. To solve this, we calculate the scale factor (K) for the vessel and use it for the vessel content and opening. However, since the loss function is translation invariant, this will lead to different translations for different objects. To solve this, we subtract the XYZ map of the content and the vessel in every pixel in which the vessel and the content overlap. The L1 distance of this property (between prediction and GT) is used as a consistency loss that promises a similar translation for all objects.

$$Translation\ Consistency\ Loss = Mean(|p_{a,i}^{vessel,GT} - p_{a,i}^{content,GT} + p_{a,i}^{vessel,Prd} - p_{a,i}^{content,Prd}|)$$

where $p_{a,i}$ is the coordinate of the point in pixel $i$ on axis $a$.

$i \in$ all the pixels in the image where the vessel and content overlap.
$a \in X, Y, Z$ axes.
*Prd* and *GT* are the predicted and ground truth, respectively.
*Vessel, Content*, refers to the type of object the point in pixel $i$ belongs to (note that only pixels where the content object and vessel object overlap are used for this loss).

## 11.3. Efficient calculation of distances and loss using dilated convolution

The difference between the XYZ coordinates for two pixels in the XYZ map can be easily calculated using a convolutional operation with a filter [1,-1]. Calculating the distance between far-away pixels can be done using dilated convolution with the distance as dilation [1,0,0,..,-1]. This enables the calculation of both the loss and the scale constant as convolutional operations, which significantly improves the running time.

## 11.4. Generating images, annotation, and depth maps using Blender

Images were created using Blender Cycles, a ray-tracing rendering tool (Figures 1, 3). Depth maps and normal maps for vessels and content were generated using Blender rendering tools and saved as .exr files. Depth maps for content were generated by simply removing the vessel from the scene, leaving its interior exposed (Figures 1, 3). The region of the vessel and content objects we're given by the vessel and content masks (Figure 1). These masks were generated by comparing the scene depth maps with and without the object and marking the regions that changed. In addition, the vessel opening plane was saved as a depth map and mask (Figure 1). This is not an actual object but identifying it is important for many applications. In order to diversify scenes, the camera position and rotation were changed randomly for each scene and are supplied in the dataset.

# 12. Acknowledgement

We acknowledge the Defense Advanced Research Projects Agency (DARPA) under the Accelerated Molecular Discovery Program under Cooperative Agreement HR00111920027, dated August 1, 2019.

The content of the information presented in this work does not necessarily reflect the position or the policy of the Government. A.A.-G. thanks Anders G. Frøseth for his generous support.

# 13. Supporting material

The TransProteus Dataset can be downloaded from:
https://www.cs.toronto.edu/matterlab/TransProteus/
https://e.pcloud.link/publink/show?code=kZfx55Zx1GOrl4aUwXDrifAHUPSt7QUAIfV
https://icedrive.net/1/6cZbP5dkNG
https://zenodo.org/record/5508261#.YUGsd3tE1H4
Code and trained models are available at:
https://github.com/sagieppel/TransProteus_Paper_Code